\documentclass[letterpaper, 10 pt, journal, twoside]{IEEEtran}
\IEEEoverridecommandlockouts                             

\usepackage{cite}

\usepackage{algorithm}
\usepackage{algpseudocode}
\usepackage{amsmath}
\usepackage{amssymb}
\usepackage{graphicx}
\usepackage{color}
\usepackage{multirow}

\newcommand{\etal}{\textit{et al. }}
\usepackage{makecell}
\usepackage{enumitem}
\setitemize{leftmargin=*}
\usepackage{balance}
\usepackage{hyperref}
\usepackage{tikz}
\usepackage{pdfprivacy}

\newcommand\copyrighttext{%
  \footnotesize \textcopyright 2022 IEEE.  Personal use of this material is permitted. Permission from IEEE must be obtained for all other uses, in any current or future media, including reprinting/republishing this material for advertising or promotional purposes, creating new collective works, for resale or redistribution to servers or lists, or reuse of any copyrighted component of this work in other works.}
\newcommand\copyrightnotice{%
\begin{tikzpicture}[remember picture,overlay]
\node[anchor=south,yshift=4pt] at (current page.south) {\parbox{\dimexpr\textwidth-\fboxsep-\fboxrule\relax}{\copyrighttext}};
\end{tikzpicture}%
}

\begin{document}

\title{\LARGE \bf Real-time Digital Double Framework to Predict Collapsible Terrains \\for Legged Robots} 
\author{
Garen Haddeler\textsuperscript{1}, Hari P. Palanivelu\textsuperscript{2}, Yung Chuen Ng\textsuperscript{2}, Fabien Colonnier\textsuperscript{2}, 
\\Albertus H. Adiwahono\textsuperscript{2}, Zhibin Li\textsuperscript{3}, Chee-Meng Chew\textsuperscript{1}, Meng Yee (Michael) Chuah\textsuperscript{2}
\thanks {$^{1}$National University of Singapore (NUS), Singapore}
\thanks {$^{2}$Institute for Infocomm Research (I\textsuperscript{2}R), A*STAR, Singapore} 
\thanks {$^{3}$Department of Computer Science, University College London, UK}
\thanks{This work is supported by the Robotics \& Autonomous Systems Department at the Institute for Infocomm Research (I\textsuperscript{2}R), Agency for Science, Technology, and Research (A*STAR), and the National University of Singapore (NUS) and Programmatic grant no. A1687b0033 from the Singapore governments Research, Innovation and Enterprise 2020 plan (Advanced Manufacturing and Engineering Domain).}
}

\maketitle
\copyrightnotice
\markboth{IEEE/RSJ INTERNATIONAL CONFERENCE ON INTELLIGENT ROBOTS AND SYSTEMS (IROS). PREPRINT VERSION. ACCEPTED JUNE 2022}%
{HADDELER \MakeLowercase{\textit{et al.}}: REAL-TIME DIGITAL DOUBLE FRAMEWORK TO PREDICT COLLAPSIBLE TERRAIN FOR LEGGED ROBOTS}

\begin{abstract}
Inspired by the digital twinning systems, a novel \emph{real-time digital double framework} is developed to enhance robot perception of the terrain conditions. Based on the very same physical model and motion control, this work exploits the use of such simulated \emph{digital double} synchronized with a \emph{real robot} to capture and extract discrepancy information between the two systems, which provides high dimensional cues in multiple physical quantities to represent differences between the modelled and the real world.
Soft, non-rigid terrains cause common failures in legged locomotion, whereby visual perception solely is insufficient in estimating such physical properties of terrains. We used \emph{digital double} to develop the estimation of the collapsibility, which addressed this issue through physical interactions during dynamic walking. 
The discrepancy in sensory measurements between the \emph{real robot} and its \emph{digital double} are used as input of a learning-based algorithm for terrain \emph{collapsibility} analysis.
Although trained only in simulation, the learned model can perform collapsibility estimation successfully in both simulation and real world. Our evaluation of results showed the generalization to different scenarios and the advantages of the \emph{digital double} to reliably detect nuances in ground conditions.
\end{abstract}
\section{Introduction}

Physics simulation has significant advancement in recent years, and has been an important tool to develop and test robot software without physical robots. 
It helps to prevent robot wear and tear, breakage, and the duplication of resources. 
Various type of simulators, capable of emulating the real world with high fidelity  %
\cite{coumans2021,gazebo,todorov2012mujoco},
have facilitated the rapid development of control algorithms and training of various control policies that can be deployed in the real world \cite{Zhao2020}.
As such simulators can be computed in real-time with appropriate hardware, we envisage a new research direction to explore the idea of digital twinning for creating new solutions in mobile robotics.

By utilizing digital twin technology, self-navigating robots can detect anomalies with robust perception and environment understanding and may take action before failing while walking on unpredictable terrain. In this work, we adopt the digital twin concept and propose a new approach by examining the information from the discrepancy between the real and the simulated systems  to develop robot perception, which is a proof of concept that extends the use of realistic physics-based simulations.

We took the advantage of embedding exactly the same control in the full physics simulation, which allows the robot to dynamically interact with the simulated environment and produce more realistic effects of foot-ground impacts. Even with the most recent advanced optimization-based planning, desired motions and forces considering impacts are very hard to model accurately or to compute in real time \cite{chatzinikolaidis2020contact}. That is why we are motivated to use the very same control and high-fidelity physics simulation, where state observations can fully capture the effects coming from more realistic physical interactions in simulated physics, e.g., impacts and their interactive effects coming from the control software. 

\begin{figure}[t]
  \centering
      \vspace{1mm} 
    \includegraphics[width=1\linewidth]{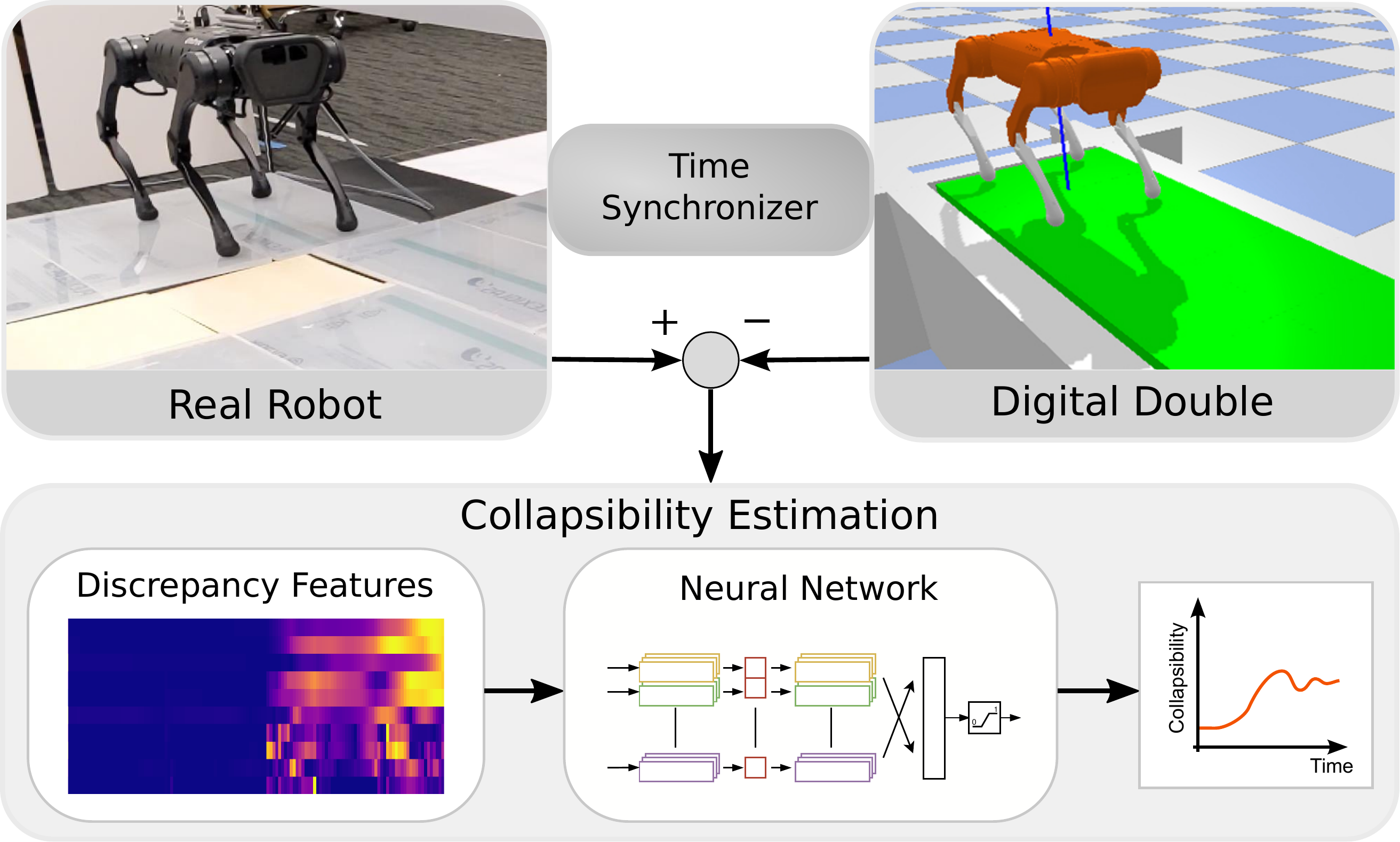}    
   \caption{Real-time digital double framework: the time synchronizer ensures a consistent timing between the \emph{real robot} and the \emph{digital double}. As the \emph{real robot} steps over an unpredictable terrain, the multi-sensory discrepancy between both robots is measured to estimate the collapsibility, and thus more reliably perceives the non-rigid ground properties during fast movements. }
    \label{fig:front_figure}
        \vspace{-6mm} 
\end{figure}

Great progress has been made in legged locomotion through optimization methods, such as blind walking on rigid and level ground \cite{bellicoso2018dynamic, kim2019highly, spotYT, yang2020multi}. 
New advancements were realized on challenging terrain by using reinforcement learning techniques \cite{kumar2021rma, takahiromiki2022} and significant improvements were also achieved in vision-aided locomotion to adjust foot placements over irregular terrains \cite{Fankhauser2018,Villarreal2019,Kim2020}.
The visual perception module can initially scan the environment during motion planning to avoid  gaps, holes, and risky obstacles.
However, the valuable information of knowing if the terrain can be rigid (concrete, asphalt, hard rock) or collapsible (mud, sand, unstable rocks) is not captured from visual feedback but only can be perceived via physical interactions.

Ensuring the dynamic stability of legged robots remains difficult while traversing areas that are collapsible, deformable, or with unstable support structures. The motivation of our work is to establish an effective detection of such anomalies for dynamic walking by examining discrepancy from simulated and real-world environments. That is, by exploiting the physical causality, we can infer the ground condition (the cause) from the resulted full-body motion (the effect). This will serve as the basis of future development on decision making and reaction strategies to these unanticipated emergencies in real-world applications.

\subsection{Related Work}
\label{sec:relatedwork}

A digital twin is defined as the digital information created by combining physical properties from the real world. 
It is a duplicate of a real-world system, and both are linked and can affect each other throughout the system's life cycle \cite{Grieves2017}. 
Digital twins can act as part of a real robot's decision making and vice-versa \cite{KRITZINGER20181016}.
In recent years, digital twin technology has been increasingly adopted in the field of robotics for decision-making in real robots by estimating information in the present and past, as well as predicting future states \cite{digitaltwin1, digitaltwin2} and monitoring operations and logistic data \cite{dtwinExample1, dtwinExample2, dtwinExample3}.

Drawing inspiration from these works, 
we developed our own framework which allows simultaneous running of a physical robot and its simulated counterpart, as well as the monitoring of the differences in sensing measurements during the synchronized execution. In this proof of concept, the simulated robot does not take part in the real robot’s decision making; thus, we hereby define the simulated robot as a \emph{digital double}.

The choice of simulator is important as it affects both the speed and fidelity of the simulation.
Based on a previous benchmark \cite{Collins2020BenchmarkingSR}, Pybullet was selected as our simulator for the \emph{digital double}.
Bullet is a lightweight physics engine that can accurately model rigid body collision, joint actuation, and non-rigid object dynamics \cite{featherstone1983}, and is well-documented and widely known in the locomotion community.

As legged robots are designed to explore areas that are inaccessible for wheeled robots, their capabilities shine in difficult terrains, such as hilly or unstable grounds that wheels cannot drive through. 
However for legged robots,
the risk of falling over when traversing unknown and unpredictable terrain lead researchers to study whether ground properties can be identified.
Mathematical models have been investigated for estimating foot-terrain interaction (hard/soft foot versus hard/soft ground) in legged robots for decades, ranging from early works by Krotkov \cite{Krotkov1990} to more recent ones by Ding \etal \cite{Ding13}. While these models make use of normal forces to analyze terrain-foot interactions, it is important to note that they have only been tested in static experimental setups, and not on actual mobile robots.

Some studies focused on static gaits and specific foot designs with embedded sensors to classify soils \cite{Kolvenbach2019} or concrete deterioration \cite{Kolvenbach2019a}. %
On hexapods, probing has also been used in combination with an adapted gait to ensure stability and possible recovery in a collapsible environment \cite{Tennakoon2020}.  Multi-sensory integration is another approach to fuse RGB-D data and joint torque measurements to classify different soils \cite{Walas2015}, where the classification technique can only be performed on previously seen similar grounds.
 The work in  \cite{Bosworth2016} showed that it was possible to estimate the ground stiffness and friction with a quadruped robot named SMC while hopping, and adapt the gait accordingly.
Cong et. al \cite{chongzhang2020} proposed a foot contact force estimation method that uses position information of the joint and applied control torque.
Wu \etal designed a capacitive tactile sensor and mounted it to the feet of a small hexapod with C-shaped rotating legs \cite{Wu2020}.
Fahmi \etal \cite{Fahmi2020} proposed soft terrain adaption and compliance estimation (STANCE) framework for their legged robot HyQ. Their proposed terrain compliance estimator registers ground reaction force, foot pose and foot velocity on a grid map and estimates stiffness and damping accordingly.

Overall, previous studies which estimate/classify terrain properties used either specially designed legged robots with additional sensors on foot \cite{Wu2020,Tennakoon2020,Kolvenbach2019a,Kolvenbach2019}, a modified static walking gait \cite{Tennakoon2020,Kolvenbach2019a,Kolvenbach2019} or a longer stance duration \cite{Fahmi2020} to capture sensory measurement accurately. 
Moreover, these studies that adopts a learning model mostly use prepared testbeds for training, thus, it may not be generalizable to classify or estimate ground properties that is not trained beforehand.
Nonetheless, these studies showed that estimation of ground characteristics is feasible within pre-trained ground characteristics, but doing so with \emph{a dynamic gait} for commercially available legged robots that solely rely on \emph{onboard sensors} (base IMU and joint sensors) to estimate unseen ground characteristics is still a challenge. 

\subsection{Contribution}
\label{sec:contribution}
Inspired by the works on twinning physical and digital systems, we introduce the ideas of using digital simulation tools and develop a new method to estimate and infer terrain properties for legged robots. 
Our contributions can be summarized as follows:
\begin{itemize}
\item The development of a novel \emph{digital double} framework with a synchronized real-time physical simulation, within which a \emph{real robot}  and its simulated physical model can use the very same control software to perform the same task simultaneously.  %
\item  Leveraging the digital double framework to extract motion discrepancy and observe the anomaly between the actual versus expected motions of the robots, where the discrepancy is formed by comparing the synchronized controls from \emph{real robot} on actual terrain and its \emph{digital double} on the ideal terrain.  
\item Effective online collapsibility estimation by correlating the motion discrepancy. The learning model is trained in simulation, and directly deployed in the real robot utilizing onboard sensors (base IMU and joint sensors), without adapting any sim-to-real techniques or retraining with real-world features.

 \end{itemize}

The remainder of the paper is organized as follows. Section~\ref{sec:methodology} presents the real-time digital double framework, defines a collapsibility metric, followed by the simulation environment, and the elaboration of using motion discrepancy as input features for designing the learning model.
Section~\ref{sec:results} reports the performance of our digital double framework in real world, the neural network input feature analysis and online collapsibility estimation in simulation and real world experiments. 
Finally, Section~\ref{sec:discussion} discusses alternative terrain property estimation approaches and advantages/disadvantages of using proposed framework,  Section~\ref{sec:conclude} concludes this work and discusses future researches.
\begin{figure}[t]
    \centering
        \vspace{2mm}

       \includegraphics[width=0.9\linewidth]{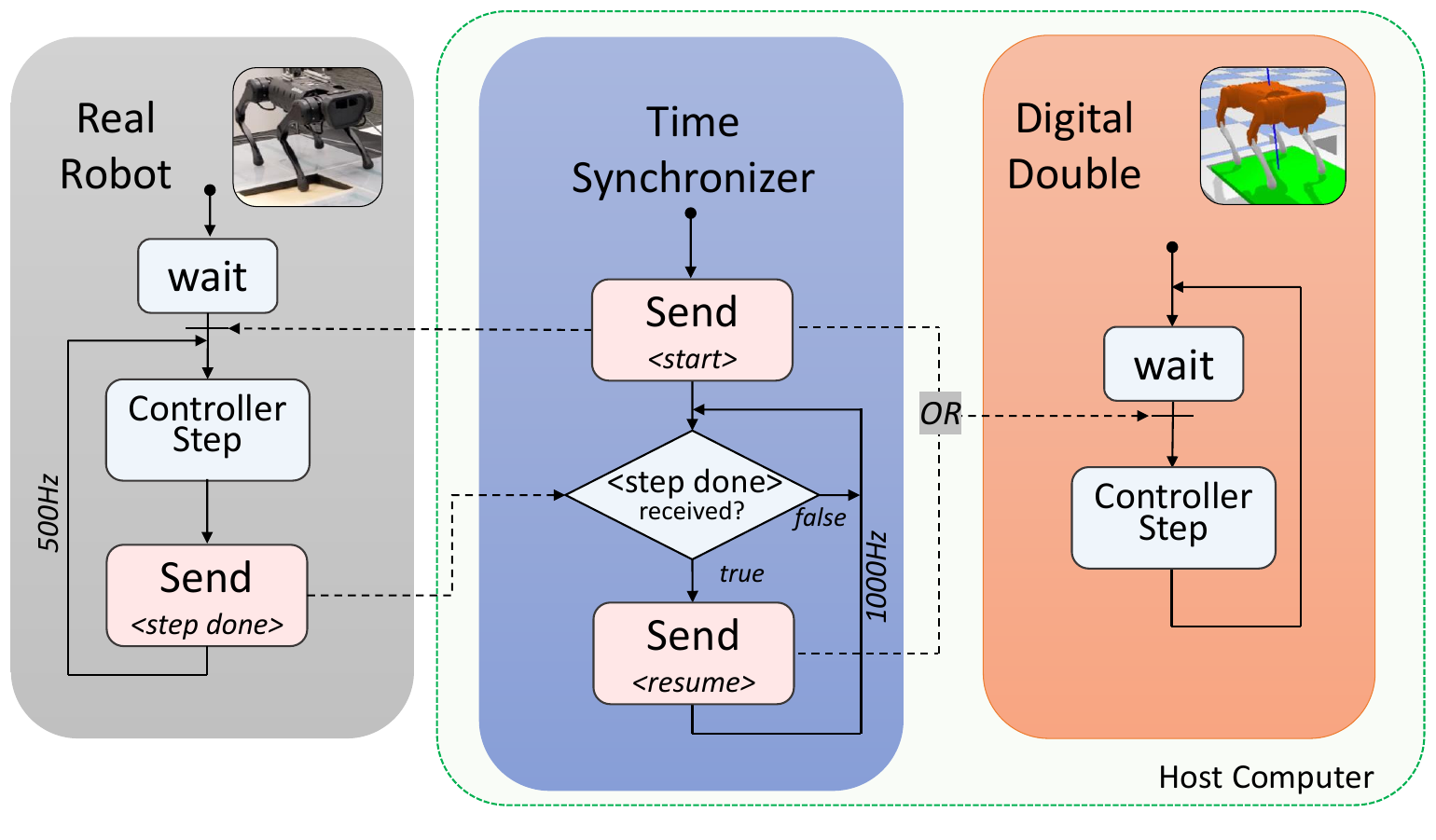}
    \vspace{-5mm}
    \caption{Time synchronizer and its interaction with the \emph{real robot} and its \emph{digital double} controllers. Same controllers are running on them and each time step is synchronized through the time synchronizer module.} 
    \label{fig:Robot_controller}
        \vspace{-5mm}
\end{figure}

\section{Methodology}
\label{sec:methodology}
In this section, the real-time digital double framework is used to measure a quantitative disturbance -- the collapsibility of the surface the robot is walking on.

\subsection{Real-time Digital Double Framework}
\label{sec:framework_setup}

The real-time digital double framework (Fig. \ref{fig:front_figure}) consists of three components: a \emph{digital double}, time synchronizer, and a \emph{real robot}. The \emph{digital double} is simulated in the PyBullet physics simulation \cite{coumans2021} in real-time, alongside the \emph{real robot}.

The real and simulated robots are based on the Unitree A1 robot \cite{unitree} with 12 actuated joints (3 motors on each leg).
To control the robot, a modular locomotion control framework from \cite{kim2019highly} is adopted which includes Model Predictive Control (MPC) and the Whole-Body Controller (WBC) methods.
Both the \emph{digital double} and the \emph{real robot} are made to run two identical but independent controllers. %

An accurate Unified Robotic Description Format (URDF) file of the Unitree A1 robot is used to reduce the modelling error between the simulation and the real world. 
The \emph{real robot} and the \emph{digital double} must both run the same control step so that they perform the same action at a given time in similar environment.
It is assumed here that the simulation runs faster than the \emph{real robot} control step. 
As such, we ensure that control loops are synced for both the \emph{real robot} and \emph{digital double}. 

To accomplish this, a time synchronizer module is formulated, as shown in Fig. \ref{fig:Robot_controller}, that synchronizes each control step loop-timings for both the real and the digital locomotion controller.
The time synchronizer module will first initialize and establish a socket connection between the \emph{real robot} and the \emph{digital double}. 
Upon initialization of the \emph{real robot} controller and \emph{digital double} controller, both wait to receive a $start$ signal from the time synchronizer in order to initialize at the same starting time.
\begin{figure}[!t]
  \centering
   \vspace{3mm}
    \includegraphics[width=\linewidth]{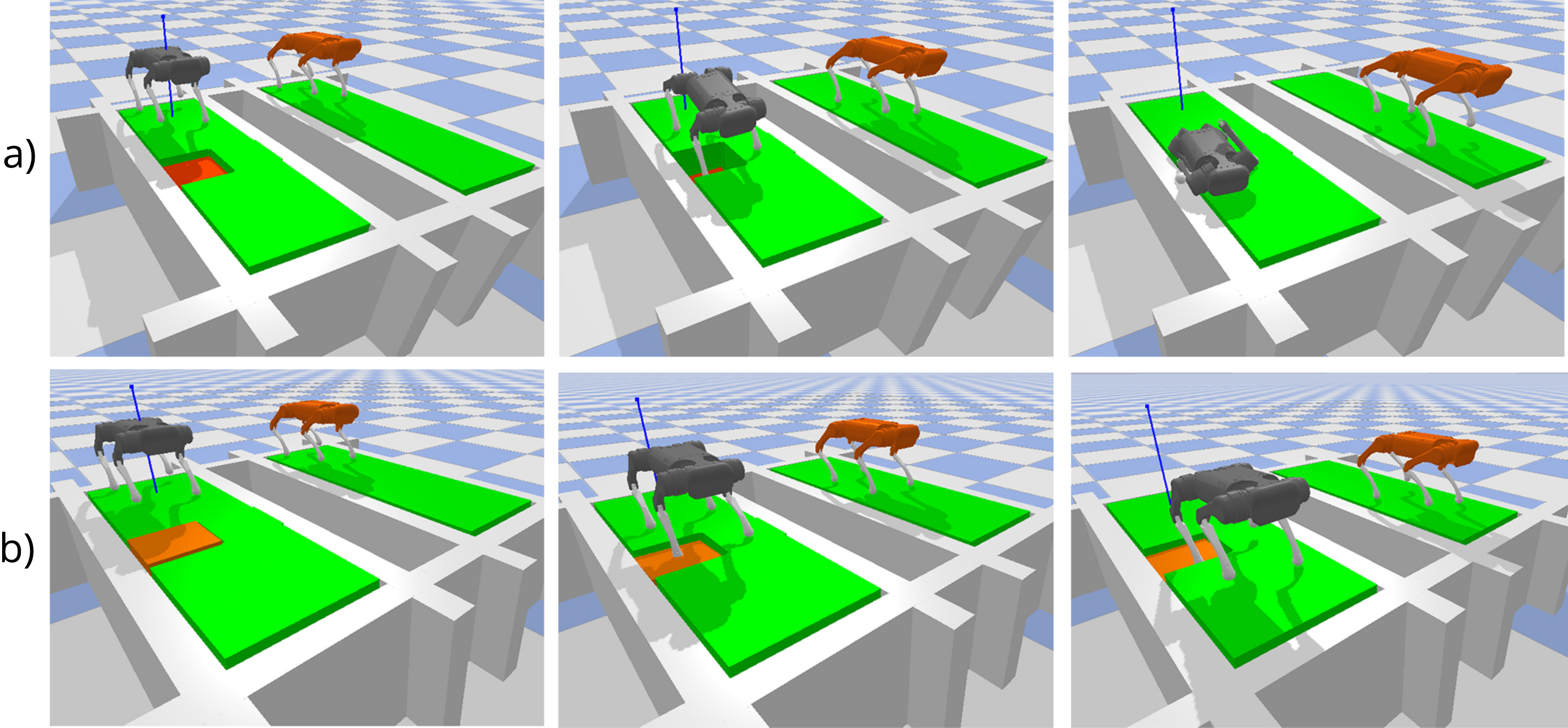} 
   \caption{Various scenarios tested in a physical simulation environment: (a) Front-Right (FR) placed Non-Rigid Tile in Collapsible case, with ground truth collapsibility of $C_{gt} \rightarrow1$ and sinking depth of $x \geq 10cm$; (b) Front-Right (FR) placed Non-Rigid Tile in Semi-Collapsible case, with $C_{gt} \in(0,1)$ and $0<x < 10cm$.}
    \label{fig:simulation_viz}
    \vspace{-5mm}
\end{figure}
After the initialization, the \emph{real robot} and the \emph{digital double} run their controller step at the same time. To ensure this, the \emph{real robot} sends the $step\text{ }done$ flag to the time-synchronizer module after each control loop step.
The message is forwarded by the time synchronizer module sending the $resume$ signal as $true$ to the digital double, which computes the next controller step. On the other hand, if the $resume$ signal is $false$, the simulated world pauses so that the \emph{real robot} can synchronize (reach the same controller step) with the \emph{digital double}.

\subsection{Collapsible Terrain}

\label{sec:Collapsibility_def}

Inspired by the work from \cite{Tennakoon2020}, we define collapsible terrain as: a terrain that looks traversable by vision, but collapses after stepping on it.
We propose to use a dimensionless definition as the ratio of the terrain vertical deformation ($x$) relative to the maximum allowed terrain deformation ($x_{max}$) that the leg can reach out to when the terrain is collapsing. Thus, $x_{max}$ is the maximum ground penetration limit that robot can withstand without failing.
\begin{equation}
    C = \frac{x}{x_{max}}
\end{equation}

The proposed collapsibility metric is used for developing anomaly detection in the context of non-rigid, collapsible surfaces using discrepancy features which are the difference between real and digital robots. Specifically, it is used to quantify (1) the deviation of foot contact and (2) learn the correspondence between non-rigid surfaces and the collapsed/sunken depths, which has a nonlinear relation, rather than identifying or discerning specific terrain properties.
Note that as long as these discrepancies are observed in the \emph{digital double}, collapsibility can be estimated on any type of terrain, and it is not limited to flat ground. 
Assuming a hole is already detected in the map, the real world and simulated environment would be reconstructed similarly. Thus, both the \emph{real robot} and the \emph{digital double} would behave similarly which would not be the case in a collapsible case.

\begin{figure*}[!t]

  \centering
      \vspace{5mm}
    \includegraphics[width=\linewidth]{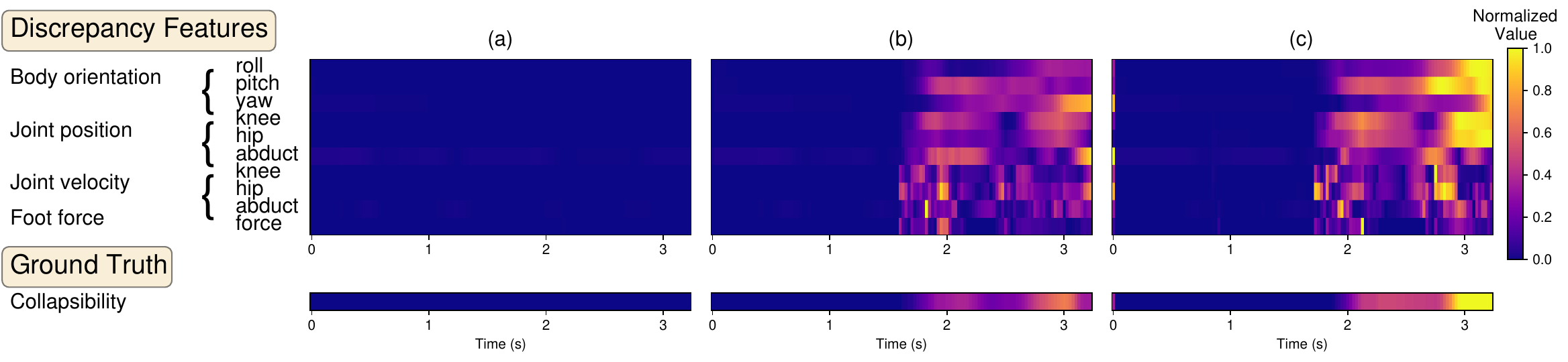}
        \vspace{-8mm}
  \caption{Front-Right leg normalized discrepancy features, \textit{i.e.} body orientation (roll, pitch, yaw), FR leg's joint position, velocity (knee, hip, abduct), and foot force errors where each feature is normalized by its maximum value. Distinguishable discrepancies appear within a short time window according to ground truth collapsibility. The instant when the robot steps into the non-rigid platform is represented as ground truth collapsibility $C_f \in [0,1]$. (a) robot walks on rigid terrain, (b) robot steps on semi-collapsible terrain with its FR leg, (c) robot steps on collapsible terrain with its FR leg. }
    \label{fig:error_image}
    \vspace{-3mm}
\end{figure*}

The collapsibility of the terrain for each foot is noted $C \in \mathbb{Q} : C \in [0,1]$. $C\approx0$ is considered as \textit{rigid} terrain, which has high stiffness and hard ground properties such that the terrain deformation is nearly $x\approx0 cm$.  \textit{Semi-collapsible} terrain  $C\in(0,1)$ has relatively less stiffness compared to rigid ground. This means that semi-collapsible ground is safer to walk on than collapsible ground while still allowing for foot movement upon ground contact. 
Therefore, the terrain deformation is in the interval $0<x<x_{max}$ where the ground has variable rigidity.
A \textit{collapsible} terrain, $C\rightarrow1$, means that the terrain exceeds allowed maximum sinking depth $x \geq x_{max}$ which will cause the robot to fall.

\subsection{Dataset Collection}

To build up the training and testing for quadrupedal walking over a collapsible platform, we developed a unique simulation environment that uses Pybullet bindings to simulate non-rigid collapsible ground \cite{coumans2021}. Two robots are spawned in a customized simulation to train the neural network model: one simulates the \emph{real robot}, and the other represents the \emph{digital double}. From the simulation, we collected both the data of sensing inputs and the ground truth values of collapsibility $C_{gt}$ to train the neural network model. 

Fig. \ref{fig:simulation_viz} shows the robot on a platform including a non-rigid tile (orange and red square), which is modeled as a spring-damper system that produces a one-dimensional displacement under the vertical external force, \textit{i.e.} contact force applied by the robot's foot. 
Thus, it allows us to simulate varying levels of vertical tile deformation. 
We placed a non-rigid square tile in 4 positions where each leg will be stepping onto the non-rigid region first respectively: Front-Left (FL), Front-Right (FR), Back-Left (BL), and Back-Right (BR). The tile's stiffness properties were adjusted to simulate  \textit{collapsible} as shown in Fig.\ref{fig:simulation_viz}-a and  \textit{semi-collapsible} cases shown as Fig. \ref{fig:simulation_viz}-b.

Four individual NN models are trained to estimate each leg's collapsibility.
The reason for an individual model for each leg is 
the disturbance of the robot’s orientation varies depending on the leg which will induce different motions to compensate during locomotion and thus induce different angular joint motion patterns.
Even though the left and right legs are symmetrical, entering with different sides into the non-rigid tile can induce different patterns.
Therefore, four models that are specifically trained for each individual leg  can capture these patterns and assess individual collapsibility for each leg. 

In total, we collected 40 trials as a configuration of FL-FR-BL-BR (4) placed tiles for semi-collapsible and collapsible cases (2), where each configuration is repeated 5 times, four for the training set and one for the test set. Duration of each semi-collapsible and collapsible trails (assuming 50Hz sampling rate) are 8 second with 400 sample and 4 second with 200 samples, respectively.

\subsection{Input for the Neural Network (NN) }

The neural network inputs are considered as the discrepancy between \emph{real robot} and \emph{digital double} measured sensory information accumulated through a time window. We formulate such discrepancy as Eq. \ref{eq:error_equation}.

\begin{equation}
\label{eq:error_equation}
    \begin{bmatrix}
    \Theta\\
    q \\
    \dot{q} \\ 
     f
    \end{bmatrix}  =     
    \begin{bmatrix} \Theta^{R} \\
   q^{R} \\
   \dot{q}^{R} \\ 
     f^{R}
 \end{bmatrix}
     -
  \begin{bmatrix} \Theta^{D} \\
    q^{D} \\
   \dot{q}^{D}  \\ 
     f^{D}
    \end{bmatrix} ,
\end{equation}
where the body orientation $\Theta$ is expressed as a vector of roll, pitch, and yaw Euler angles, while $q$ and $ \dot{q}$ are positions and velocities, respectively, for knee, hip and abduct joints. Finally, $f$ represents feet forces.
These features are defined by the difference between the \emph{real robot}  and \emph{digital double} and are denoted with superscripts $R$ and $D$, respectively.

The discrepancy feature image in Fig. \ref{fig:error_image} visualizes body orientation error (roll, pitch, yaw), FR leg's joint position, velocity (knee, hip, abduct), and foot force error where each feature is normalized by its maximum value.
Robot steps over FR placed tiles when \textit{non-collapsible} (Fig. \ref{fig:error_image}-a), \textit{semi-collapsible} (Fig. \ref{fig:error_image}-b) and \textit{collapsible} (Fig. \ref{fig:error_image}-c) tile in simulation.
Distinguishable patterns can be observed from the image whereas the discrepancy increases in proportion to the ground truth collapsibility.

Indeed, the discrepancy arises relatively to how collapsible the surface is, \textit{i.e.} when the robot steps on a surface, exerting smaller reaction forces due to the collapse of the ground, there is no physically available or sufficient reaction forces to reduce the position errors. When the robot enters a collapsible area, the discrepancy significantly increases between the real measurements and their expected values from the \emph{digital double}. As a result, the subsequent errors during the stance phase rise and persist, despite any feedback control. 
This discrepancy and level of deviation from the nominal, expected motions, therefore, inherit recognizable patterns for the estimation of the collapsibility.

\begin{figure}[!t]
    \vspace{2mm}
    \centering
    \includegraphics[width=0.95\linewidth]{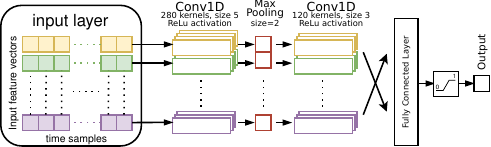}
    \caption{The neural network structure for collapsibility estimation. }
    \label{fig:NN_architecture}
    \vspace{-4mm}
\end{figure}

\subsection {Neural Network (NN) Design}
\label{ssec:NN_design}
The proposed neural network represented as Fig.\ref{fig:NN_architecture} is composed of two 1D-convolution layers, each followed by a ReLu activation layer with a Max pool layer in between. Then, 2 fully-connected layers are used to provide a single value output which is collapsibility estimation. The 1D-convolutions are independent for each variable. The two convolution layers output 280 and 120 channels with a kernel size of 5 and 3, respectively. This structure was selected as the convolution layers are able to capture temporal patterns among the data. In total, there are 243641 trainable parameters when 9 input features are used. This structure was selected as the convolution layers are able to capture temporal patterns among the data.

Input features are accumulated over a time window of $0.5$ second, where the sliding window time is chosen as,$0.04$ second. We have observed that the selected time window size is enough to capture features in one gait cycle. Different sets of input features were investigated through several training runs. For each training, 100 epochs are processed with a learning rate kept constant at 0.001.

 \begin{figure}[t]
  \centering
        \vspace{2mm}

        \includegraphics[width=1\linewidth]{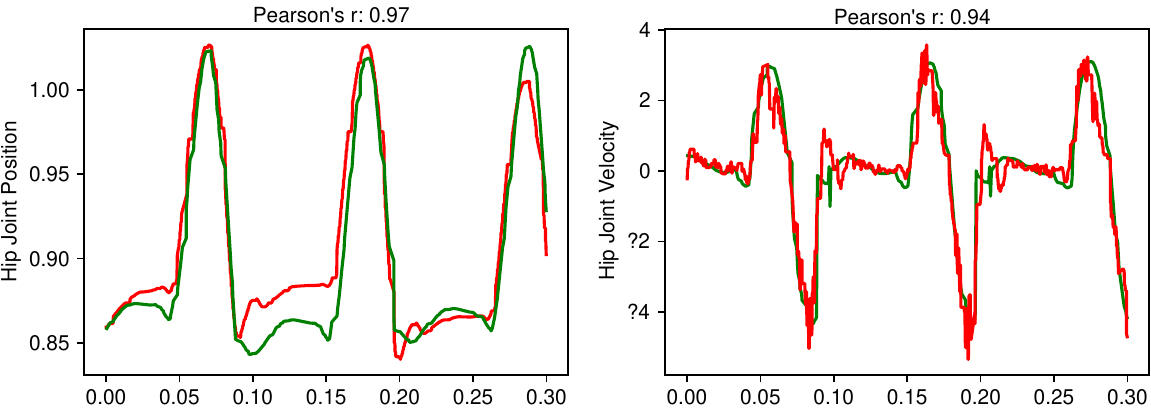}

          \vspace{-4mm}

  \caption{Performance of the real-time digital double framework: a-b) the position and velocity profiles of the FR hip joint, from the \emph{real robot} and its \emph{digital double} respectively during trotting on a flat ground. The high Pearson's r score shows a strong correlation of two signals from the \emph{real robot} and its \emph{digital double}. }
      \label{fig:close_joint}
          \vspace{-5mm}
\end{figure}

\section{Results}
\label{sec:results}

We present our results in 5 subsections. First, an assessment of real-time digital double framework in real-world experiment is presented. Thereafter, an analysis of the NN parameters reveals the influence of the input features, and the NN's performance is assessed on the test dataset for each leg. Finally, the online collapsible estimation in simulation and real-world experiments is presented for semi-collapsible and collapsible cases.

\subsection{Assessment of the Real-time Digital Double Framework}

In order to demonstrate the performance of the real-time digital double framework, the \emph{real robot} and its \emph{digital double} walked on a flat rigid surface.
The robot's on-board computer does the computations for the locomotion controller, while the time synchronizer and physics simulation runs on a host PC linked to the robot via Ethernet. 
For the host PC, we use an Intel Core i7-10870H CPU with 8 cores, each with a clocking speed of 2.20 GHz, with a GeForce RTX 3070 GPU for simulation rendering. 

Without the time synchronizer, the \emph{digital double}'s control loop finishes faster ($1109 \mu s$) than the real one ($2203 \mu s$).
Such large difference in average control loop time causes large motion discrepancy between two systems since both controllers does not perform the same action at given time. 
As a result, we observed non-periodic time delays in the scale of milliseconds on both robots' control loop instead of periodic patterns due to occasional time delay in computation and in data transfer. 
With the implementation of the time synchronizer, both \emph{real robot} and \emph{digital double} compute their control loop with an average time difference of $9 \mu s$, thus both controllers perform similar action at a given time.
Fig. \ref{fig:close_joint} shows the close look-up of the FR hip joint position and velocity for the \emph{real robot} and the \emph{digital double}.
We can observe that the \emph{digital double} has similar dynamic results to the \emph{real robot} under standard settings (assuming robots walking level ground) since both joint position and velocities have matching amplitudes. Within the time window of 1 seconds, phases of the \emph{real robot} and \emph{digital double} are fully synchronized.  
A strong positive correlation on joint position and velocity is observed with Pearson’s $r$ scores of $0.97$ and $0.95$ respectively, where the Pearson's $r$ score in statistics is defined as a linear correlation coefficient of two sets of data. 
\begin{table}[!t]

\centering
\small{
\scriptsize
\vspace{3mm}
\caption{Test Errors by Input Feature}
\begin{tabular}{|l|l|l|l|}
	\hline
	Order & Symbol & Discrepancy Feature Definition & Size \\ \hline
	\multirow{2}{*}{Zero}&  $\Theta$ & Body Orientation Angle& 3\\

	                            & $q$  & Joint Position* & 3 \\
\hline 
		\multirow{1}{*}{First} & $\dot{q}$ & Joint Velocity*&3  \\
	\hline
			\multirow{1}{*}{Higher} &  $f$  & Foot Force* \tiny{  *in a single leg}& 1 \\
	\hline

\end{tabular}\\
\vspace{0.3em}
\centering
\scriptsize
\begin{tabular}{|l|l|l|}
\hline
\makecell{Input \\ Feature} &Error($\%$)& Variance \\\hline
$\Theta$   & 9.06&0.012\\ \hline
$q$ &  5.32&0.006 \\ \hline
$\Theta$,$q$ & 5.56&0.006 \\\hline
$\dot{q}$ &4.23&0.005 \\\hline
$f$ & 5.195&0.008\\\hline
\end{tabular}
\begin{tabular}{|l|l|l|}
\hline
\makecell{Input \\ Feature}& Error ($\%$) & Variance  \\ \hline
$\dot{q}$,$f$  &  7.02& 0.011 \\ \hline
$q$,$\dot{q}$  &  4.19& 0.003\\ \hline
$q$,$f$ &  6.62 &0.012 \\ \hline
\textbf{$\Theta$,$q$,$\dot{q}$} & \textbf{3.817}& \textbf{0.003}\\ \hline
$\Theta$,$q$,$\dot{q}$,$f$ & 5.121 & 0.004 \\ \hline
\end{tabular}
}
\label{tab:Training_acc_variables}
\vspace{-4mm}
\end{table}

\subsection{Analysis of NN Inputs}
\label{sec:impact_of_variables}

 All the quantities involved in the physical interactions can potentially provide correlations between their discrepancy and the level of collapsibility. Our analysis explores how informative each variable is, and our results show  which discrepancy features reflect and correlate more with an abnormal gait over the collapsible terrains.  

Inputs for the model are grouped by the order of derivatives (zero, first, higher) in Table \ref{tab:Training_acc_variables}, and the NN is trained with different combinations to quantify the contribution of each variable.
The  mean absolute error that is used to evaluate performance of the models is calculated as $MAE(\%) = 100\frac{1}{k}\sum_{t=0}^k|C_{gt}(t) - \hat{C}(t)|$, $|C_{gt}(t) - \hat{C}(t)| \in [0,1]$ by first calculating the absolute difference of estimation ($\hat{C}$) and ground truth collapsibility ($C_{gt}$) at time stamp $t$ and then averaging it over time $k$. 
In order to ensure reliable statistics in analysis of NN inputs, the error mean and  its variance is calculated from five different training cycles. 
From Table \ref{tab:Training_acc_variables}, according to our analysis, we can observe that solely using body orientation error ($\Theta$) gives the highest error, however when used in combination with joint position and joint velocity (set $\Theta$, $q$, $\dot{q}$) achieves the lowest error.

Unlike the related works presented in Section \ref{sec:relatedwork}, our study shows that contact force information $f$ alone does not provide the best results and degrades the performance when in combination with others  (shown in Table \ref{tab:Training_acc_variables}). 
 Using contact force information for collapsibility estimation can be feasible in static walking or active probing motion, but in dynamic walking, impact force induces noises and spikes in small time-window since the robot locomotion controller balances the torso and legs in a compliant and dynamic manner rather than standing and only probing the ground.
Thus, for our learning pipeline the level of noises and spikes in force data decrease the quality of the collapsibility estimation on our test dataset. 
In this regard, for our proposed deep learning model, the collapsibility is more easily predicted using the zero or first-order features as inputs -- they form the principle and primary components for online estimation of collapsibility. 
By studying all the combinations, we finalized  the set of features ($\Theta$, $q$ and, $\dot{q}$) as the input for our proposed deep learning model, as it renders the highest accuracy.

Moreover, according to our statistical observation on the testing set retrieved during the dataset collection, individual leg's NN models have  $R^2=[0.87,0.9,0.91,0.94]$ values which indicate a high level of correlation between the estimated collapsibility and the ground truth. 
As a result, we can conclude that all four models are capable of correlating the motion discrepancy features with the continuous ground truth collapsibility encountered during training.

 \begin{figure}[t]

  \centering
    \vspace{2mm}
    \includegraphics[width=1\linewidth]{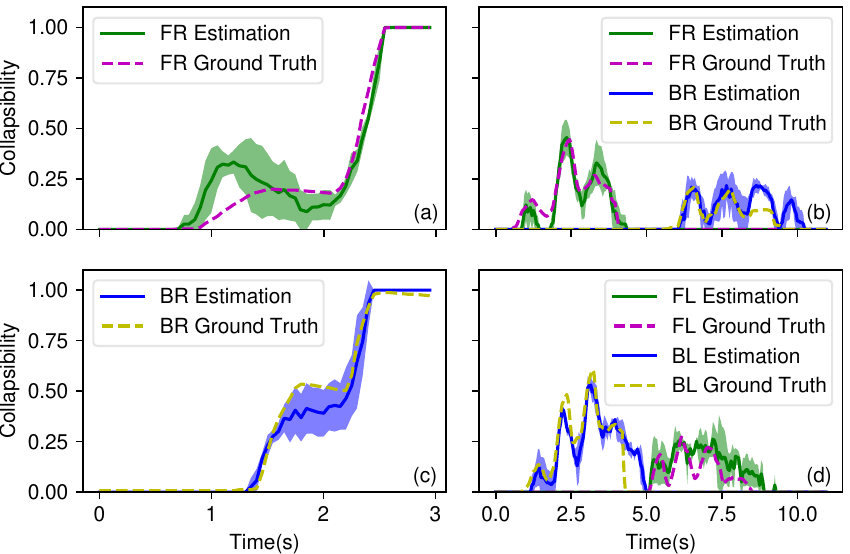}

    \vspace{-4mm}
  \caption{Online collapsibility estimation in the Simulation: (a)
  Front-Right (FR) foot estimation, where a collapsible tile was placed FR and its sinking depth was more than $x_{max}$ ($x\geq10cm$, $C_{gt} \rightarrow 1$) ; (b) FR and Back-Right (BR) foot estimation, where a semi-collapsible tile with $C_{gt} \in(0,1)$ was placed at FR; (c) BR foot estimation where a collapsible tile was placed at BR; (d) Front-Left (FL) and Back-Left (BL) foot estimation where a semi-collapsible tile was placed at BL.}

    \label{fig:testing_col_and_semi_col}
    \vspace{-4mm}
\end{figure}
\subsection{Collapsible Terrain Estimation in Simulation}
Real-time online collapsibility estimation in simulation is presented to demonstrate the performance of the trained model.
Multiple scenarios are carried out to test the capabilities of an individual foot's collapsibility estimation.
Four of the outcomes that have similar ground truth plots are used to generate the mean collapsibility estimation and its upper-lower boundaries.

Fig. \ref{fig:simulation_viz}(a) displays FR foot stepping on collapsible tile ($x\geq10cm$, $C_{gt} \rightarrow 1$) placed in the front-right corner and Fig. \ref{fig:testing_col_and_semi_col} (a) shows its estimation graph.
The same collapsible tile is also placed back-right and the BR feet estimation graph is presented as Fig. \ref{fig:testing_col_and_semi_col} (c) when the robot is walking backward.

Fig. \ref{fig:simulation_viz}(b) shows FR and BR feet stepping on semi-collapsible tile ($0<x<10cm$, $C_{gt} \in(0,1)$) with both front and back legs in sequence as the robot continues walking forward and
its online estimation result is shown as Fig.\ref{fig:testing_col_and_semi_col} (b).
Similarly, semi-collapsible tile placed back-left and while the robot is walking backwards, the FL, BL feet estimation graph is captured as in Fig. \ref{fig:testing_col_and_semi_col} (d).
We can observe that the FR-FL-BR-BL feet models are capable of precisely estimating non-collapsible, semi-collapsible, and collapsible tiles as well as identifying them separately.

\subsection{Real world Implementation of Collapsibility Estimation }
\label{sec:real_world_implementation}

\begin{figure}[t]
  \centering
    \vspace{2mm}
       \includegraphics[width=1\linewidth]{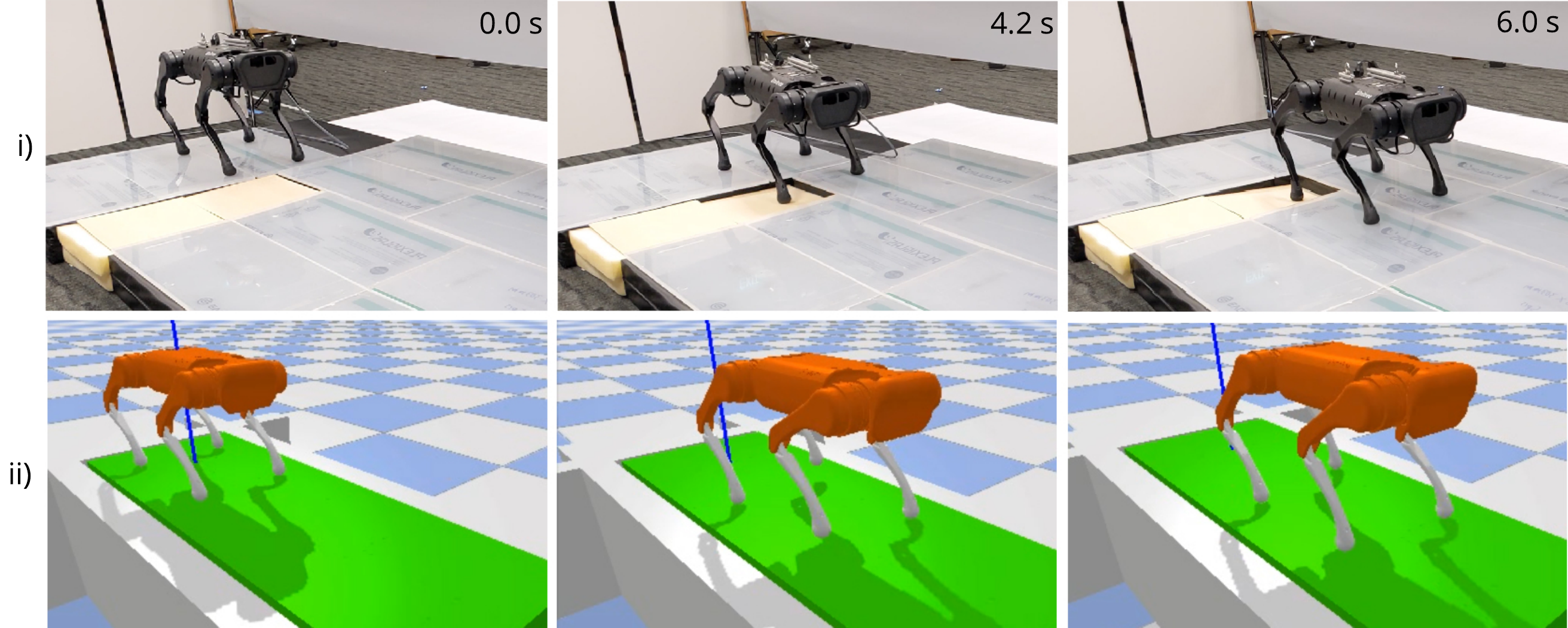}

  \caption{Snapshots of real world experiments with the \emph{digital double} running in parallel: (i) the \emph{real robot} steps over front-right placed semi-collapsible tile, where the real situation differs from the expectation; (ii) the \emph{digital double} simultaneously walks on a hard ground as if in an ideal scenario. Time is stamped according to result of semi-collapsible case  shown in Fig. \ref{fig:real_world_result}(b-i) and in Fig. \ref{fig:real_world_result}(b-ii)}
   \label{fig:sim_real_snashot}
    \vspace{-5mm}
\end{figure}

In the experimental setting shown in Fig. \ref{fig:sim_real_snashot}(i), \emph{real robot} walks on semi-collapsible terrain while \emph{digital double} walks on hard ground  in Fig. \ref{fig:sim_real_snashot}(ii).
Unitree A1 quadruped robot \cite{unitree} is used along with the host computer that simulates \emph{digital double} and executes the time synchronizer.

Collapsibility ground truth ($C_{gt}$) is calculated through the foot's relative vertical displacement and the foot position with respect to a fixed frame is obtained from a real-time 2D motion analysis software -- Kinovea. 
Thereafter, accuracy is calculated by averaging the absolute difference of estimation and ground truth over time.

\begin{figure}[!t]
  \centering
      \vspace{2mm}

    \includegraphics[width=1\linewidth]{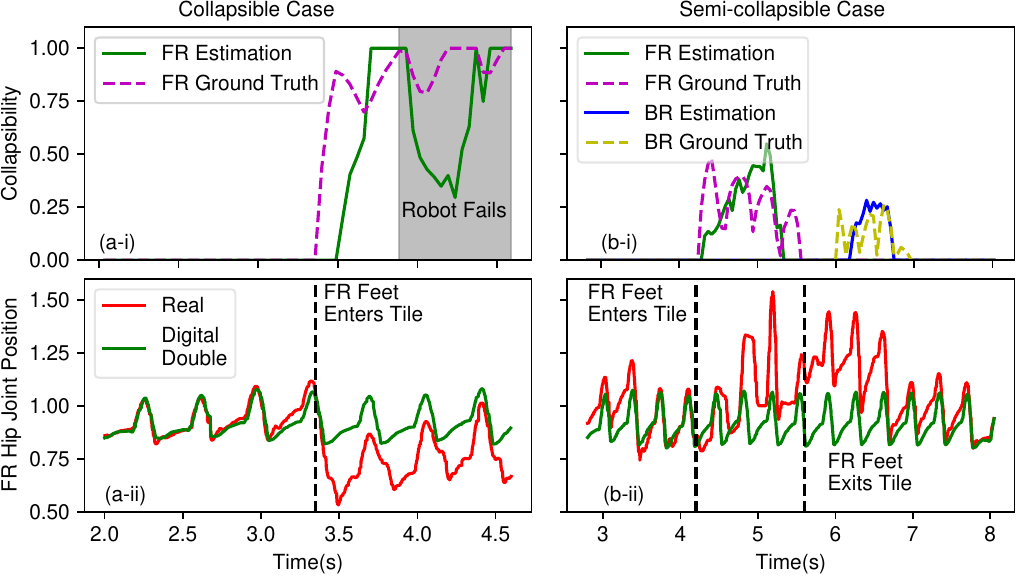}
        \vspace{-4mm}
  \caption{Online collapsibility estimation in Real-world Environment: (a-i) Front-Right (FR) feet estimation, where a collapsible tile was placed FR and its sinking depth is more than $x_{max}$ ($x\geq10cm$, $ C_{gt} \rightarrow 1$); (b-i) FR and Back-Right (BR) feet estimation, where a semi-collapsible tile with $C_{gt}\in(0,1)$; (a-ii and b-ii) Front-Right hip's joint Position in real and \emph{digital double} while FR foot enters the non-rigid tile.}
    \label{fig:real_world_result}
    \vspace{-6mm}
\end{figure}

Fig. \ref{fig:real_world_result}(a-i) shows FR placed collapsible tile estimation where non-rigid tile can deform such that ground truth $C_{gt}\rightarrow1$, sinking foot depth becomes $x\geq10cm$.
A discrepancy between \emph{real robot} and \emph{digital double} can be seen from the hip joint positions when FR foot steps on the collapsible tile in Fig. \ref{fig:real_world_result}(a-ii).
FR foot successfully estimates collapsibility with $6.7\%$ of MAE compared to ground truth during the experiment.
After the robot falls ($3.8s$ onward), estimation fails since joint failures was not part of the training process.

Fig. \ref{fig:real_world_result}(b-i) shows FR placed semi-collapsible tile estimation where non-rigid tile can deform such that ground truth $C_{gt}\in[0,1]$ and sinking foot depth becomes $0<x<10cm$.
Therefore, the robot could cross over it with its FR and BR legs and both is able to estimate collapsible tile with  $3.43$\%, $1.23\%$ of MAE compared to ground truth during the experiment.
The motion FR hip joint discrepancy between the \emph{real robot} and \emph{digital double} can be seen in Fig. \ref{fig:real_world_result}(b-ii).
While the BR foot is still on the non-rigid tile, a FR hip joint discrepancy can still be seen when the FR foot leaves the tile.
However, the FR collapsibility estimator appropriately estimates as non-collapsible ($\hat{C}\approx0$) after the FR foot leaves the non-rigid tile.
 This is the benefit of using four individual NN model since each model can capture patterns for each foot's actual collapsibility even though there is discrepancy on one of the input features.

It should be highlighted that the estimation was trained in simulation and then directly deployed on the \emph{real robot}, with no additional retraining using real data. This shows a clear advantage of using the proposed \emph{digital double} to learn how to infer the ground property.

\section{Discussion}
\label{sec:discussion}

For static walking, the terrain stiffness, as the changes in forces versus that in spatial displacements ($\frac{ \Delta F}{\Delta x}$), can possibly detect the property of contact surfaces. However, during \textit{dynamic walking}, it is more challenging for several reasons.
Firstly, IMU and joint encoders have limited resolutions and have phase lag/mismatch between them, which can cause significant errors in the end-effector positions. Thus, the precision of the calculated spatial position of the foot is difficult during fast locomotion.  
Secondly, adding a force sensor at foot for directly sensing reduces stability during dynamic walking, due to increased mass-inertia. Also, estimating force indirectly via joint torques is not accurate during dynamic motions. Finally, in dynamic walking, the stepping frequency can be 2 Hz or higher, with a very short stance time of 0.16-0.25 s. During the stance, contact/impact forces are largely fluctuating and spiking, making stiffness calculations very problematic in such a short time window. 

As alternative approach for motion discrepancy,  \emph{real robot}'s sensory information solely can be used to train the learning model and estimate collapsibility. 
Moreover, rather than using digital double framework, motion discrepancy could have been estimated if the \emph{real robot}'s motion pattern can be captured offline and synchronized with the \emph{real robot} online. 
The main drawback of these methods is that the dataset collection must account for all possible robot actions and scenarios, including varying speeds, directions, gaits, and more. As a result, collecting and training all conceivable combinations is inefficient and time-consuming.
Thus, one advantage of using a real-time digital double is the scalability. It can effortlessly deliver motion discrepancy in different robots and payload mass without pre-training or pre-recording since the robot's model and controller can be easily updated in a simulated world.
We decided to use motion discrepancies between reality and expectations, so that the learning model can be more generalizable and perform effectively in untrained real-world cases which are tested in Section \ref{sec:real_world_implementation}. 

One of the major disadvantage of utilizing a \emph{digital double} for terrain estimate is the assumption that the simulation world has similar geometric ground properties to the real world. In reality, rough terrain might have irregular and uneven shapes, making the \emph{digital double} technique less effective.
In our proof of concept, we show in Section \ref{sec:real_world_implementation} that if the simulated environment have similar terrain shapes and location to the real one, we can have a similar estimation performance. As we will demonstrate in future work,  by incorporating SLAM, we can recreate perceived real-world geometric shapes in the physics simulation while also running the proposed framework.
 
The advantage of simulating more realistic interactions comes at a higher computational cost. However, according to our experiment, our framework uses only 7.5\% of overall computing in the PC (Intel Core i7-10870H CPU with 8 cores and  2.20 GHz clocking speed). Although we used a host PC to simulate \emph{digital double} in this study, it is feasible to run whole framework in the robot onboard computer.

\section{Conclusion}
\label{sec:conclude}
We presented the proof of concept of a real-time digital double framework for legged robots to extend the robot's sensing on real-world terrains. By utilizing the time synchronizer on two independent locomotion control loops, \emph{digital double} and the \emph{real robot} were able to behave similarly in real-time. 
The motion discrepancy, between the \emph{digital double} walking in hard ground versus \emph{real robot} walking on non-rigid terrain surfaces, is used for online detection of anomalies correlated to collapsibility of the terrain. The estimation interpolates semi-collapsible grounds as relatively safe to step on, compared to the collapsible ground. %

We benchmarked the relative influence of each input feature in the neural network model and concluded the effective combination is to use the discrepancy of joint positions, joint velocities, and body orientation, which are easily available in most platforms and provide good results in collapsibility estimation. We showed that it is feasible to analyze the collapsibility of the ground in dynamic legged motion with \textit{minimal latency} and high accuracy using the proposed method.
No retraining or data collection in real world was necessary, even though the training was solely done in simulation, which is an advantage of using high-dimensional sensory date based on the digital double framework.

Future work will investigate uneven and outdoor cases to test proposed collapsibility estimation framework. Moreover, our method can be used for the research in reactive locomotion on quadruped robots. Fully integrated digital twin, as in the literature, can also be investigated with vision feedback for mitigation upon detected undesired behaviors.

\balance
\bibliographystyle{IEEEtran}
\bibliography{bib/bibliography}

\begin{thebibliography}{10}
\providecommand{\url}[1]{#1}
\csname url@samestyle\endcsname
\providecommand{\newblock}{\relax}
\providecommand{\bibinfo}[2]{#2}
\providecommand{\BIBentrySTDinterwordspacing}{\spaceskip=0pt\relax}
\providecommand{\BIBentryALTinterwordstretchfactor}{4}
\providecommand{\BIBentryALTinterwordspacing}{\spaceskip=\fontdimen2\font plus
\BIBentryALTinterwordstretchfactor\fontdimen3\font minus
  \fontdimen4\font\relax}
\providecommand{\BIBforeignlanguage}[2]{{%
\expandafter\ifx\csname l@#1\endcsname\relax
\typeout{** WARNING: IEEEtran.bst: No hyphenation pattern has been}%
\typeout{** loaded for the language `#1'. Using the pattern for}%
\typeout{** the default language instead.}%
\else
\language=\csname l@#1\endcsname
\fi
#2}}
\providecommand{\BIBdecl}{\relax}
\BIBdecl

\bibitem{coumans2021}
\BIBentryALTinterwordspacing
E.~Coumans and Y.~Bai, ``Pybullet, a python module for physics simulation for
  games, robotics and machine learning,'' 2016-2022. [Online]. Available:
  \url{http://pybullet.org}
\BIBentrySTDinterwordspacing

\bibitem{gazebo}
N.~Koenig and A.~Howard, ``Design and use paradigms for gazebo, an open-source
  multi-robot simulator,'' in \emph{IEEE/RSJ International Conference on
  Intelligent Robots and Systems}, Sendai, Japan, Sep 2004, pp. 2149--2154.

\bibitem{todorov2012mujoco}
E.~Todorov, T.~Erez, and Y.~Tassa, ``Mujoco: A physics engine for model-based
  control,'' in \emph{2012 IEEE/RSJ international conference on intelligent
  robots and systems}.\hskip 1em plus 0.5em minus 0.4em\relax IEEE, 2012, pp.
  5026--5033.

\bibitem{Zhao2020}
W.~Zhao, J.~P. Queralta, and T.~Westerlund, ``Sim-to-real transfer in deep
  reinforcement learning for robotics: a survey,'' in \emph{2020 IEEE Symposium
  Series on Computational Intelligence (SSCI)}, 2020, pp. 737--744.

\bibitem{chatzinikolaidis2020contact}
I.~Chatzinikolaidis, Y.~You, and Z.~Li, ``Contact-implicit trajectory
  optimization using an analytically solvable contact model for locomotion on
  variable ground,'' \emph{IEEE Robotics and Automation Letters}, vol.~5,
  no.~4, pp. 6357--6364, 2020.

\bibitem{bellicoso2018dynamic}
C.~D. Bellicoso, F.~Jenelten, C.~Gehring, and M.~Hutter, ``Dynamic locomotion
  through online nonlinear motion optimization for quadrupedal robots,''
  \emph{IEEE Robotics and Automation Letters}, vol.~3, no.~3, pp. 2261--2268,
  2018.

\bibitem{kim2019highly}
D.~Kim, J.~Di~Carlo, B.~Katz, G.~Bledt, and S.~Kim, ``Highly dynamic quadruped
  locomotion via whole-body impulse control and model predictive control,''
  \emph{arXiv preprint arXiv:1909.06586}, 2019.

\bibitem{spotYT}
\BIBentryALTinterwordspacing
{Boston Dynamics}. With you, spot can. [Online]. Available:
  \url{https://youtu.be/VRm7oRCTkjE}
\BIBentrySTDinterwordspacing

\bibitem{yang2020multi}
C.~Yang, K.~Yuan, Q.~Zhu, W.~Yu, and Z.~Li, ``Multi-expert learning of adaptive
  legged locomotion,'' \emph{Science Robotics}, vol.~5, no.~49, 2020.

\bibitem{kumar2021rma}
A.~Kumar, Z.~Fu, D.~Pathak, and J.~Malik, ``Rma: Rapid motor adaptation for
  legged robots,'' \emph{arXiv preprint arXiv:2107.04034}, 2021.

\bibitem{takahiromiki2022}
T.~Miki, J.~Lee, J.~Hwangbo, L.~Wellhausen, V.~Koltun, and M.~Hutter,
  ``Learning robust perceptive locomotion for quadrupedal robots in the wild,''
  \emph{Science Robotics}, vol.~7, no.~62, Jan 2022.

\bibitem{Fankhauser2018}
P.~Fankhauser, M.~Bjelonic, C.~D. Bellicoso, T.~Miki, and M.~Hutter, ``{Robust
  Rough-Terrain Locomotion with a Quadrupedal Robot},'' \emph{Proceedings -
  IEEE International Conference on Robotics and Automation}, pp. 5761--5768,
  2018.

\bibitem{Villarreal2019}
O.~A.~V. Magaña, V.~Barasuol, M.~Camurri, L.~Franceschi, M.~Focchi, M.~Pontil,
  D.~G. Caldwell, and C.~Semini, ``Fast and continuous foothold adaptation for
  dynamic locomotion through cnns,'' \emph{IEEE Robotics and Automation
  Letters}, vol.~4, no.~2, pp. 2140--2147, 2019.

\bibitem{Kim2020}
D.~Kim, D.~Carballo, J.~{Di Carlo}, B.~Katz, G.~Bledt, B.~Lim, and S.~Kim,
  ``{Vision Aided Dynamic Exploration of Unstructured Terrain with a
  Small-Scale Quadruped Robot},'' \emph{Proceedings - IEEE International
  Conference on Robotics and Automation}, pp. 2464--2470, 2020.

\bibitem{Grieves2017}
M.~Grieves and J.~Vickers, \emph{{Digital Twin : Mitigating Unpredictable ,
  Undesirable Emergent Behavior in Complex Systems}}.\hskip 1em plus 0.5em
  minus 0.4em\relax Springer International Publishing, 2017, pp. 85--113.

\bibitem{KRITZINGER20181016}
W.~Kritzinger, M.~Karner, G.~Traar, J.~Henjes, and W.~Sihn, ``Digital twin in
  manufacturing: A categorical literature review and classification,''
  \emph{IFAC-PapersOnLine}, vol.~51, no.~11, pp. 1016--1022, 2018, 16th IFAC
  Symposium on Information Control Problems in Manufacturing INCOM 2018.

\bibitem{digitaltwin1}
P.~Staczek, J.~Pizon, W.~Danilczuk, and A.~Gola, ``A digital twin approach for
  the improvement of an autonomous mobile robot (amr’s) operating
  environment-a case study,'' \emph{Sensors}, vol.~21, no.~23, 2021.

\bibitem{digitaltwin2}
J.~Douthwaite, B.~Lesage, M.~Gleirscher, R.~Calinescu, J.~M. Aitken,
  R.~Alexander, and J.~Law, ``A modular digital twinning framework for safety
  assurance of collaborative robotics,'' \emph{Frontiers in Robotics and AI},
  vol.~8, 2021.

\bibitem{dtwinExample1}
M.~Grzelak, A.~Borucka, and A.~Świderski, ``Assessment of the influence of
  selected factors on the punctuality of an urban transport fleet,''
  \emph{Transport Problems}, vol.~15, pp. 311--323, 12 2020.

\bibitem{dtwinExample2}
G.~Bocewicz, I.~Nielsen, A.~Gola, and Z.~Banaszak, ``Reference model of
  milk-run traffic systems prototyping,'' \emph{International Journal of
  Production Research}, vol.~59, pp. 1--18, 06 2020.

\bibitem{dtwinExample3}
P.~Sitek, J.~Wikarek, K.~Rutczynska-Wdowiak, G.~Bocewicz, and Z.~Banaszak,
  ``Optimization of capacitated vehicle routing problem with alternative
  delivery, pick-up and time windows: a modified hybrid approach,''
  \emph{Neurocomputing}, vol. 423, 05 2020.

\bibitem{Collins2020BenchmarkingSR}
J.~J. Collins, J.~B. McVicar, D.~Wedlock, R.~Brown, D.~Howard, and J.~Leitner,
  ``Benchmarking simulated robotic manipulation through a real world dataset,''
  \emph{IEEE Robotics and Automation Letters}, vol.~5, pp. 250--257, 2020.

\bibitem{featherstone1983}
R.~Featherstone, ``The calculation of robot dynamics using articulated-body
  inertias,'' \emph{The international journal of robotics research}, vol.~2,
  no.~1, pp. 13--30, 1983.

\bibitem{Krotkov1990}
E.~Krotkov, ``Active perception for legged locomotion: every step is an
  experiment,'' in \emph{Proceedings. 5th IEEE International Symposium on
  Intelligent Control 1990}, 1990, pp. 227--232 vol.1.

\bibitem{Ding13}
L.~Ding, H.~Gao, Z.~Deng, J.~Song, Y.~Liu, G.~Liu, and K.~Iagnemma,
  ``Foot–terrain interaction mechanics for legged robots: Modeling and
  experimental validation,'' \emph{The International Journal of Robotics
  Research}, vol.~32, no.~13, pp. 1585--1606, 2013.

\bibitem{Kolvenbach2019}
H.~Kolvenbach, C.~Bartschi, L.~Wellhausen, R.~Grandia, and M.~Hutter, ``{Haptic
  inspection of planetary soils with legged robots},'' \emph{IEEE Robotics and
  Automation Letters}, vol.~4, no.~2, pp. 1626--1632, 2019.

\bibitem{Kolvenbach2019a}
H.~Kolvenbach, G.~Valsecchi, R.~Grandia, A.~Ruiz, F.~Jenelten, and M.~Hutter,
  ``{Tactile Inspection of Concrete Deterioration in Sewers with Legged
  Robots},'' \emph{Field and Service Robots (FSR)}, no. August, p.~14, 2019.

\bibitem{Tennakoon2020}
E.~Tennakoon, T.~Peynot, J.~Roberts, and N.~Kottege, ``{Probe-before-step
  walking strategy for multi-legged robots on terrain with risk of collapse},''
  \emph{Proceedings - IEEE International Conference on Robotics and
  Automation}, pp. 5530--5536, 2020.

\bibitem{Walas2015}
K.~Walas, ``{Terrain Classification and Negotiation with a Walking Robot},''
  \emph{Journal of Intelligent and Robotic Systems: Theory and Applications},
  vol.~78, no. 3-4, pp. 401--423, 2015.

\bibitem{Bosworth2016}
W.~Bosworth, J.~Whitney, S.~Kim, and N.~Hogan, ``Robot locomotion on hard and
  soft ground: Measuring stability and ground properties in-situ,'' in
  \emph{2016 IEEE International Conference on Robotics and Automation (ICRA)},
  2016.

\bibitem{chongzhang2020}
Z.~Cong, A.~Honglei, C.~Wu, L.~Lang, Q.~Wei, and M.~Hongxu, ``Contact force
  estimation method of legged-robot and its application in impedance control,''
  \emph{IEEE Access}, vol.~8, pp. 161\,175--161\,187, 2020.

\bibitem{Wu2020}
X.~A. Wu, T.~M. Huh, A.~Sabin, S.~A. Suresh, and M.~R. Cutkosky, ``Tactile
  sensing and terrain-based gait control for small legged robots,'' \emph{IEEE
  Transactions on Robotics}, vol.~36, no.~1, pp. 15--27, 2020.

\bibitem{Fahmi2020}
S.~Fahmi, M.~Focchi, A.~Radulescu, G.~Fink, V.~Barasuol, and C.~Semini,
  ``Stance: Locomotion adaptation over soft terrain,'' \emph{IEEE Transactions
  on Robotics}, vol.~36, no.~2, pp. 443--457, 2020.

\bibitem{unitree}
\BIBentryALTinterwordspacing
X.~Wang. Unitree robotics. [Online]. Available: \url{http://www.unitree.com}
\BIBentrySTDinterwordspacing

\end{thebibliography}

\end{document}